\documentclass{llncs}
\usepackage[T1]{fontenc}
\usepackage[colorlinks=true, linkcolor=blue, citecolor=blue, urlcolor=blue]{hyperref}
\usepackage{graphicx}
\usepackage{booktabs}
\usepackage{tabularx}
\usepackage{placeins}
\usepackage{amsmath}
\usepackage{orcidlink}
\usepackage{fancyhdr}

\begin{document}
\title{Personalizing Mathematical Game-based Learning for Children: A Preliminary Study}
\titlerunning{Personalizing Mathematical Game-based Learning}
%
%
\author{Jie Gao\orcidlink{0000-0002-6933-950X} \and
Adam K. Dubé \orcidlink{0000-0001-9935-8886}}
\authorrunning{J. Gao and A. K. Dubé}
%
\institute{McGill University, Montreal, Quebec, Canada\\
\email{{jie.gao3}@mail.mcgill.ca, {adam.dube}@mcgill.ca}}
\maketitle              
\begin{abstract}
Game-based learning (GBL) is widely adopted in mathematics education. It enhances learners’ engagement and critical thinking throughout the mathematics learning process. However, enabling players to learn intrinsically through mathematical games still presents challenges. In particular, effective GBL systems require dozens of high-quality game levels and mechanisms to deliver them to appropriate players in a way that matches their learning abilities. To address this challenge, we propose a framework, guided by adaptive learning theory, that uses artificial intelligence (AI) techniques to build a classifier for player-generated levels. We collect 206 distinct game levels created by both experts and advanced players in Creative Mode, a new tool in a math game-based learning app, and develop a classifier to extract game features and predict valid game levels. The preliminary results show that the Random Forest model is the optimal classifier among the four machine learning classification models (k-nearest neighbors, decision trees, support vector machines, and random forests). This study provides insights into the development of GBL systems, highlighting the potential of integrating AI into the game-level design process to provide more personalized game levels for players.

\keywords{Personalize \and Game-based Learning \and  Machine Learning \and Mathematics.}
\end{abstract}

%
%

\section{Introduction}
Game-based learning (GBL) has been widely examined for its effectiveness and affordance in enhancing students’ critical thinking, problem-solving, and conceptual understanding~\cite{alam2025systematic,dai2022narrative,ke2016designing,liljedahl2016problem}. The recent meta-analysis research shows that GBL has a significant positive overall effect on students' critical thinking~\cite{mao2022effects}. These findings align with previous research and also highlight the critical role of specific game mechanics~\cite{kacmaz2022examining}. A good math game goes beyond gamification and turns the core math skill into game mechanics (i.e., what the player does) to make learning the game~\cite{sharma2022game}. Ke~\cite{ke2016designing} called these intrinsically integrated educational games and showed they are far more effective than games where learning is extrinsic to gameplay (i.e., play the game then solve math problems). 

Making intrinsic games requires dedicated designers and hours of labor for each level, while mastering math skills requires cumulative practice over dozens of levels~\cite{mayfield2002effects}. However, most GBL systems face this challenge~\cite{dube2016games}. Konca~\cite{konca2022digital} evaluated 30 popular STEM applications and showed that most math applications paid more attention to content quality, design, functionality, and technical features. Some researchers also indicated that both intrinsic motivation for math and the quality of the playing experience impact students' learning success~\cite{ninaus2017acceptance}.

To make the game levels more intrinsically mathematical and to improve students’ learning of mathematical knowledge, a math game-based learning app developed a tool, Creative Mode, that helps players enhance their creative thinking skills. Through this tool, players can design new game levels and expert-selected user levels can be delivered to other players. Yet, manually identifying which player-levels are suitable for sharing with other players is untenable, thousands of levels have to be reviewed by the experts. Additionally, it is challenging to provide adaptive game levels that match individuals’ learning abilities. Previous research focuses on using machine learning (ML) methods to predict learner performance. However, there is limited research on filtering player-generated game levels using ML models~\cite{chi2014choice,lee2024comparison}.
To address this concern, we aimed to design a framework that (1) uses a machine learning classifier to identify and classify valid player-generated levels and (2) provide personalized game levels to appropriate players, all guided by adaptive learning theory.

\section{Related Work}
\subsection{GBL in Mathematics} 
GBL is defined as a learning method that integrates digital games into the learning environment~\cite{chang2020experiencing}. Some researchers suggested that well-designed math games can provide multi-level behavioural, cognitive, and affective interactions that increase children’s interest and performance in math \cite{kacmaz2022examining,mcewen2015engaging}. Previous studies have indicated that GBL can address pedagogical concerns by providing low-stakes, student-centered contexts where abstract laws and concepts are concretized during gameplay~\cite{sharma2022game}, improve students' meaning-making abilities when interacting with multiple external representations framed by the GBL mechanics~\cite{pan2023patterns}, and provide adaptive challenges matching students' learning abilities~\cite{chiotaki2023adaptive,sharma2022game}.

An increasing body of evidence has indicated that GBL improves not only students' learning outcomes but also their critical thinking skills~\cite{cicchino2015using,mao2022effects}. By simulating real-world problems in a learning environment, students can try different strategies over time and adjust their solutions based on the feedback; thereby improving their critical thinking~\cite{mao2022effects}. Furthermore, moving beyond the role of a player, the transition to a game level creator encourages students to engage in higher-order thinking. In this role, they are tasked with the complex challenge of constructing innovative levels that effectively balance entertainment value with pedagogical efficacy~\cite{hsiao2014development}.

\subsection{Math Game Adaptivity} 
Previous research~\cite{sharma2022game} suggests that games can be designed to adapt to students' actions and level of learning, achieving game adaptivity. However, Cayton-Hodges and her colleagues~\cite{cayton2015tablet} reviewed 64 math apps on four dimensions, including mathematical content, feedback, scaffolding, interactions, and adaptability. They found that not all apps focused on adaptability. Some researchers indicated that most educational games today implement a low-resolution adaptivity form at the level of individual players~\cite{vandewaetere2013adaptivity}. They argue that a micro-adaptive, dynamic, fine-grained, and player-centered approach to game design is critical.

\subsection{Current Study} 
The present study focuses on exploring an effective approach to identifying valid play-generated game levels and providing personalized levels to individual players, better fulfilling their learning needs. The purpose of this work is to provide insights into how artificial intelligence (AI) techniques can be integrated into GBL for math learning. It also addresses a critical question, which is whether intrinsic educational games can be designed to provide sufficient practice for students to achieve mastery. 

\section{Methodology}

\begin{figure}[t!]
    \centering
    \includegraphics[width=\columnwidth]{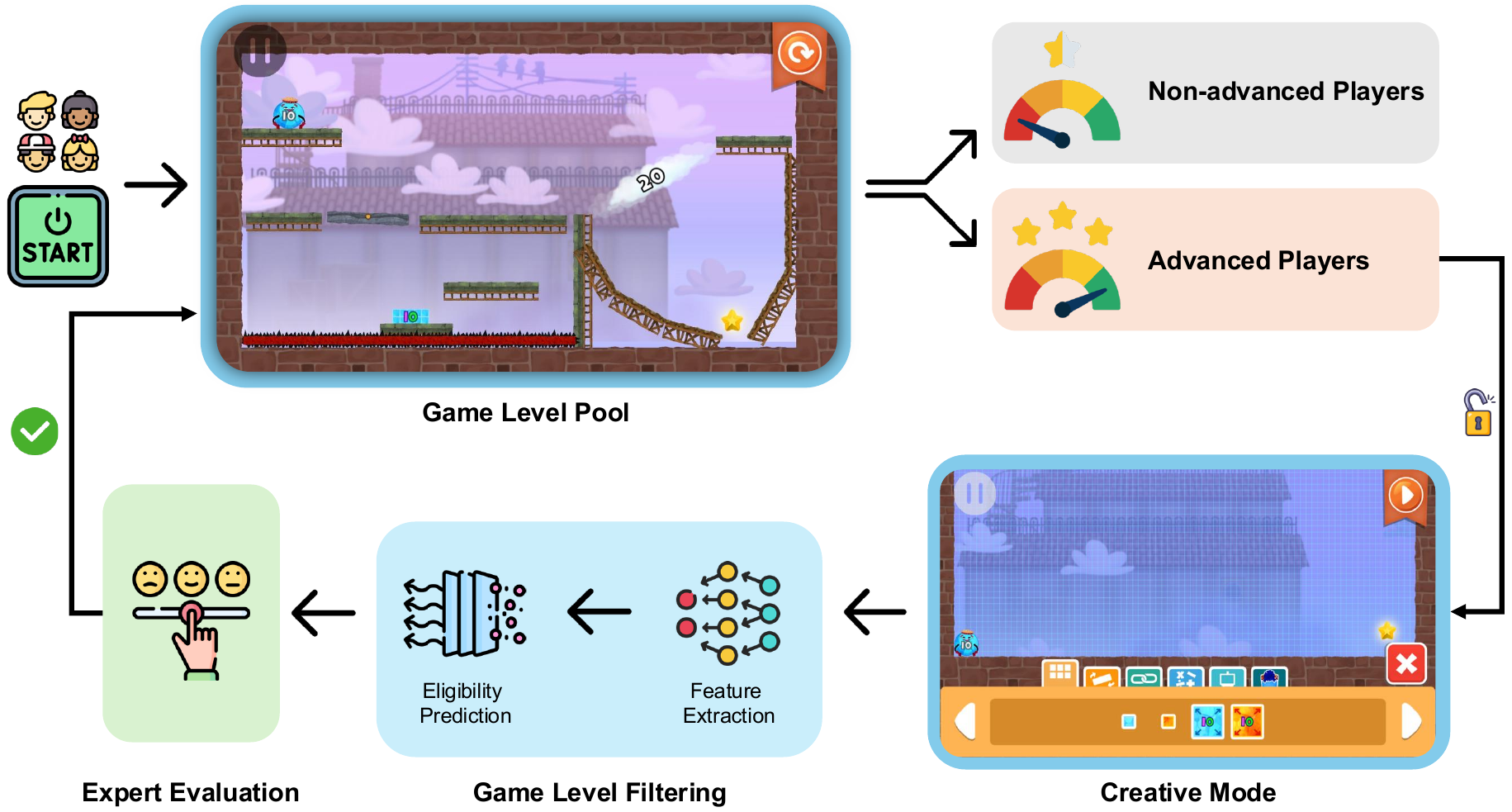}
    \caption{Overview of the player-generated levels selection pipeline.}
    \vspace{0pt}
\label{p}
\end{figure}

\subsection{Dataset}
In this study, we employ a game level dataset from a math game-based learning app, gathered between 2024 and 2025. This app builds conceptual understanding through game-based learning, supporting children ages 5-10 around the world. Creative Mode is a tool within the game, allowing advanced players to design and create new game levels based on their own ideas. This mode helps advanced players improve their creative thinking with math and provides the cumulative practice that other players need through player-generated levels. 

The dataset includes 206 distinct game levels from Creative Mode: 86 expert-designed levels and 120 player-generated levels. Each level is associated with a JSON file that encodes all game features. Among player-generated levels, 44 of them were validated and selected by experts at the game company for other peers to play,  while the remaining levels were excluded from the active pool. We extract key variables for this study, including the counts of features (N=25) and their associated values (N=12). Note that some features appear more than once within a level. The dataset is fully anonymous.

\vspace{1em}
\begin{table}[!t]
\centering
\caption{Four variable groups: player character, goal, physics objects, and obstacles.}
\label{t}

\begin{tabularx}{\linewidth}{
  >{\raggedright\arraybackslash}p{0.2\linewidth}
  >{\raggedright\arraybackslash}X
  >{\raggedright\arraybackslash}X
}
\toprule
\textbf{Groups} & \textbf{Variable Examples} & \textbf{Level Examples} \\
\midrule

Player Character
&
Player Character, Player Character Value
&
\begin{minipage}[t]{\linewidth}
\textbf{}\\[-2pt]
  \includegraphics[width=0.9\linewidth]{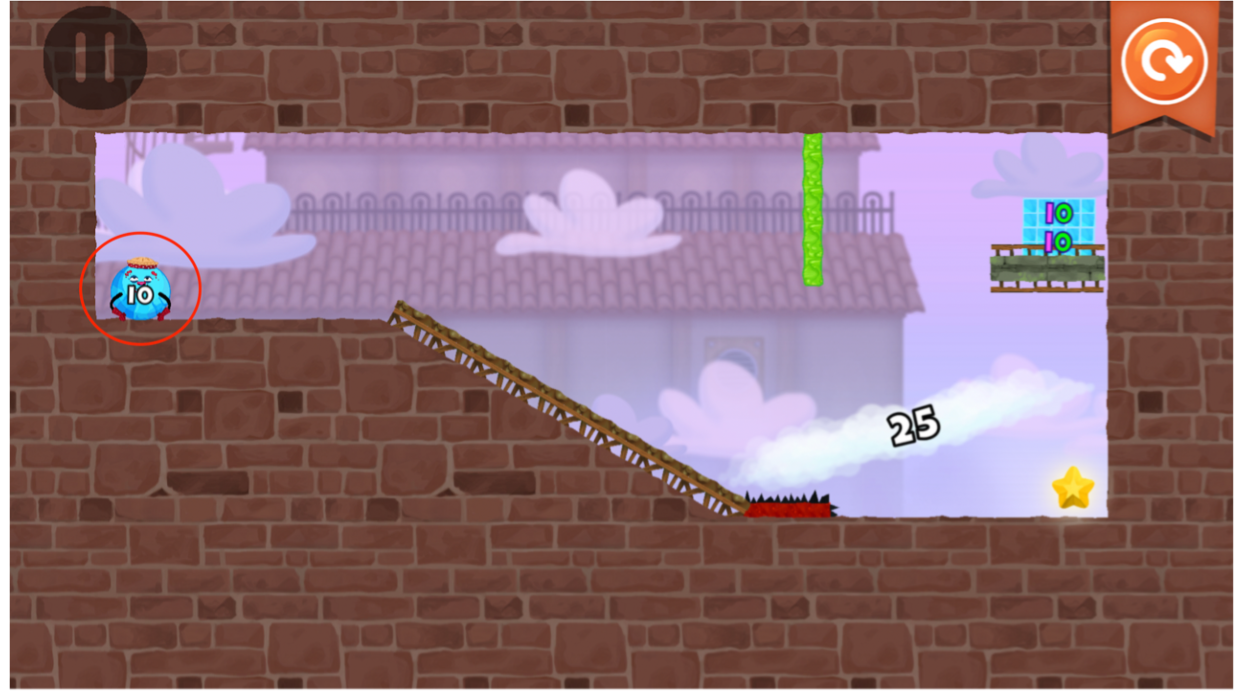}
\end{minipage}
\\
\midrule

Goal
&
Goal
&
\begin{minipage}[t]{\linewidth}
\textbf{}\\[-2pt]
  \includegraphics[width=0.9\linewidth]{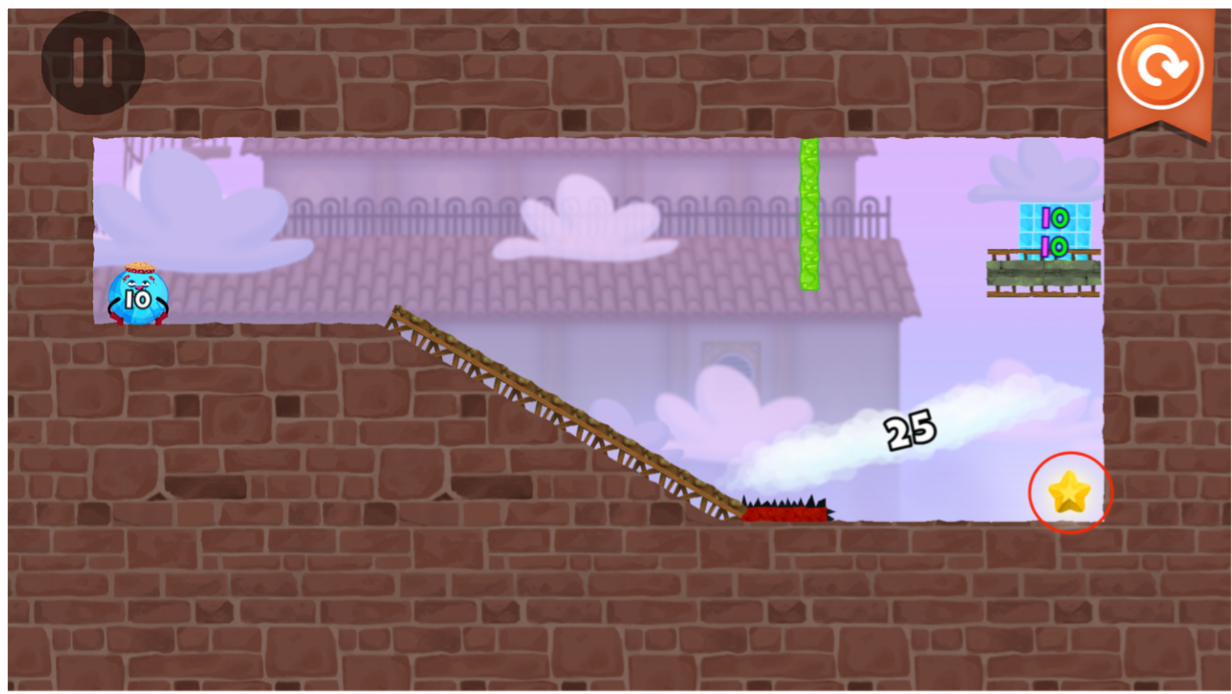}
\end{minipage}
\\
\midrule

Physics Objects
&
Ice Block x1, Ice Block x 10, Lava Block x1, Lava Block x10, Bubble, One-way Platform, Slimy Platform, Sticky Platform
&
\begin{minipage}[t]{\linewidth}
\textbf{}\\[-2pt]
  \includegraphics[width=0.9\linewidth]{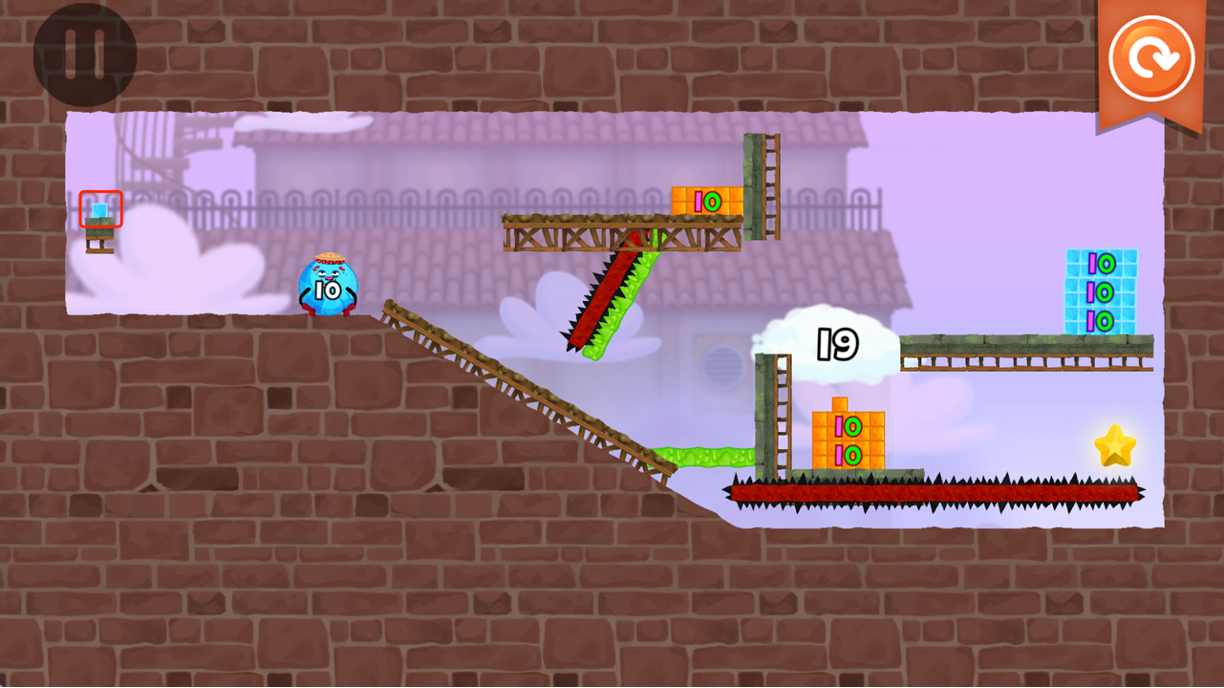}
\end{minipage}
\\
\midrule

Obstacles
&
Cloud, Could Value, Door, Door Value, Spiky Platform, Breakable Wall, Breakable Wall Value
&
\begin{minipage}[t]{\linewidth}
\textbf{}\\[-2pt]
  \includegraphics[width=0.9\linewidth]{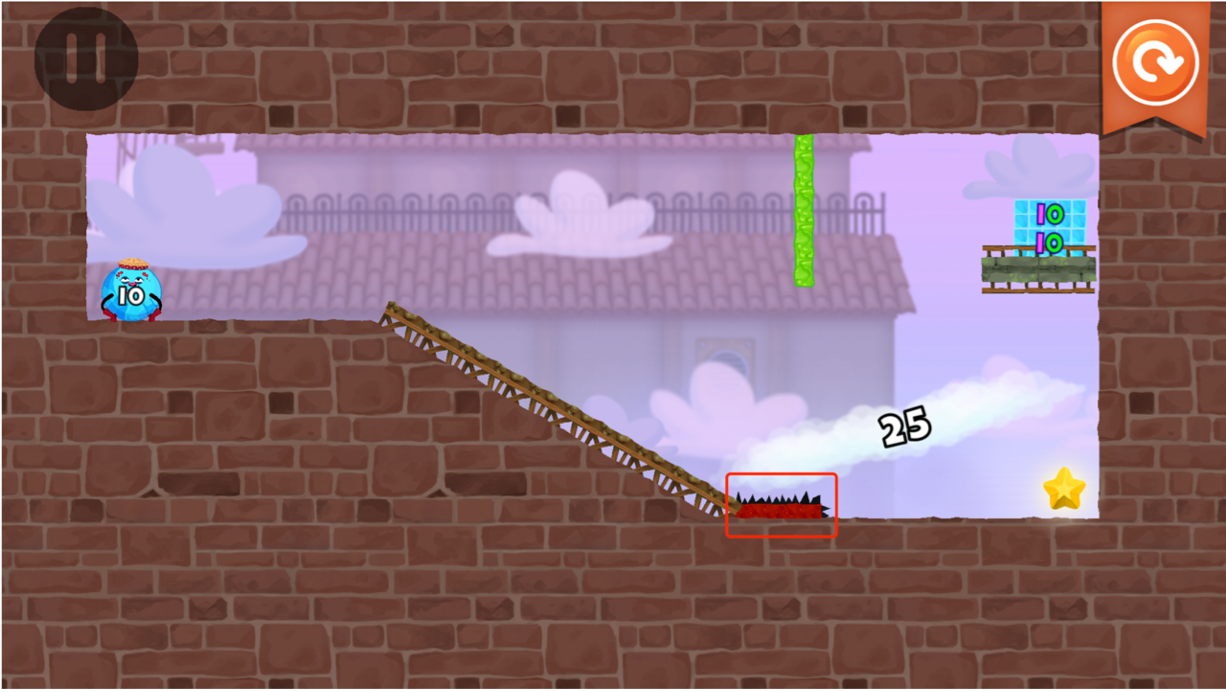}
\end{minipage}
\\

\bottomrule
\end{tabularx}
\end{table}

\subsection{Design Pipeline}
As illustrated in Figure~\ref{p}, advanced players gain access to the Creative Mode and can drag elements (e.g., ice blocks, lava blocks, bubbles, and platforms) into the board to design a game level. After creating the game levels, we extract the key variables from each level and apply a machine learning approach to filter the levels. Selected levels identified by the model are sent to experts for a final review. Only validated levels are approved and deployed into the public game pool for other players to play.

\subsection{Data Processing and Analysis}
The data processing mainly consists of the following steps. First, we extracted key variable values from each level and stored the data in a structured dataset. Second, we classified the 61 extracted variables into four groups: player character, goal, physics objects, and obstacles (see Table~\ref{t}). The player character group includes variables related to the game character. The goal group represents the completion goal of the game. Each game level includes a single goal. The physics objects group comprises diverse game objects, including ice blocks, lava blocks, and platforms. Some of these objects have individual values. When the character approaches such an object, the character's value merges (e.g., addition, subtraction) with the object's value. The obstacles group includes game objects that may prevent the character from moving, such as clouds, doors, and breakable walls. Third, missing values were recoded as missing. Fourth, we used Lasso, a regression analysis method, to further select features and reduce the dimensionality of the data, enhancing prediction accuracy and result interpretability.

This study applied four ML models to identify the optimal predictive model: k-nearest neighbors (KNN), decision trees (DT), support vector machines (SVM), and random forests (RF). We selected these algorithms because they represent a diverse range of classification mechanisms. Moreover, model performance is assessed using nested cross-validation, which can minimize evaluation bias.

\section{Preliminary Results}

This section presents the results of the model performance metrics of four selected ML models. The metrics include accuracy, precision, recall, F1-score, ROC-AUC score, and the confusion matrix. The results are preliminary, mainly focusing on the phase of selecting the most effective model as the game level classifier.

\vspace{1em}
\begin{table}[!t]
\centering
\caption{Results of the classification performance evaluation (Inner vs. Outer loop).}
\label{t1}

{\fontsize{6}{10}\selectfont

\begin{tabularx}{\linewidth}{
  >{\raggedright\arraybackslash}p{0.1\linewidth}
  >{\centering\arraybackslash}p{0.17\linewidth}
  >{\centering\arraybackslash}p{0.17\linewidth}
  >{\centering\arraybackslash}p{0.17\linewidth}
  >{\centering\arraybackslash}p{0.17\linewidth}
  >{\centering\arraybackslash}p{0.17\linewidth}
}
\toprule
\textbf{Model} & \textbf{Accuracy} & \textbf{Precision} & \textbf{Recall} & \textbf{F1-Score} & \textbf{ROC-AUC} \\
\midrule

\multicolumn{6}{l}{\textbf{Inner loop}} \\
\midrule
KNN & \mbox{81.42 $\pm$ 1.53\%} & \mbox{73.85 $\pm$ 3.92\%} & \mbox{65.32 $\pm$ 3.13\%} & \mbox{66.41 $\pm$ 3.53\%} & \mbox{84.77 $\pm$ 4.71\%} \\
DT  & \mbox{83.89 $\pm$ 1.54\%} & \mbox{78.06 $\pm$ 3.57\%} & \mbox{73.40 $\pm$ 4.72\%} & \mbox{73.63 $\pm$ 3.96\%} & \mbox{81.58 $\pm$ 3.66\%} \\
SVM & \mbox{82.42 $\pm$ 1.19\%} & \mbox{75.90 $\pm$ 3.14\%} & \mbox{66.67 $\pm$ 3.23\%} & \mbox{68.11 $\pm$ 3.64\%} & \mbox{\textbf{86.84 $\pm$ 2.20\%}} \\
RF  & \mbox{82.07 $\pm$ 1.37\%} & \mbox{74.43 $\pm$ 2.64\%} & \mbox{70.46 $\pm$ 2.18\%} & \mbox{71.12 $\pm$ 2.27\%} & \mbox{85.86 $\pm$ 1.85\%} \\
\midrule

\multicolumn{6}{l}{\textbf{Outer loop}} \\
\midrule
KNN & \mbox{79.90 $\pm$ 4.01\%} & \mbox{68.60 $\pm$ 10.79\%} & \mbox{62.97 $\pm$ 8.19\%} & \mbox{63.66 $\pm$ 9.10\%} & \mbox{84.12 $\pm$ 8.45\%} \\
DT  & \mbox{82.23 $\pm$ 3.88\%} & \mbox{75.45 $\pm$ 7.54\%} & \mbox{71.72 $\pm$ 7.39\%} & \mbox{71.86 $\pm$ 6.60\%} & \mbox{81.69 $\pm$ 7.62\%} \\
SVM & \mbox{81.93 $\pm$ 2.53\%} & \mbox{74.90 $\pm$ 6.93\%} & \mbox{65.52 $\pm$ 4.98\%} & \mbox{67.50 $\pm$ 5.19\%} & \mbox{86.22 $\pm$ 4.87\%} \\
RF  & \mbox{82.42 $\pm$ 5.71\%} & \mbox{74.53 $\pm$ 8.91\%} & \mbox{72.69 $\pm$ 9.75\%} & \mbox{72.70 $\pm$ 9.22\%} & \mbox{\textbf{86.57 $\pm$ 6.32\%}} \\
\bottomrule
\end{tabularx}
}
\end{table}


\begin{figure}[t!]
    \centering
    \includegraphics[width=0.6\columnwidth]{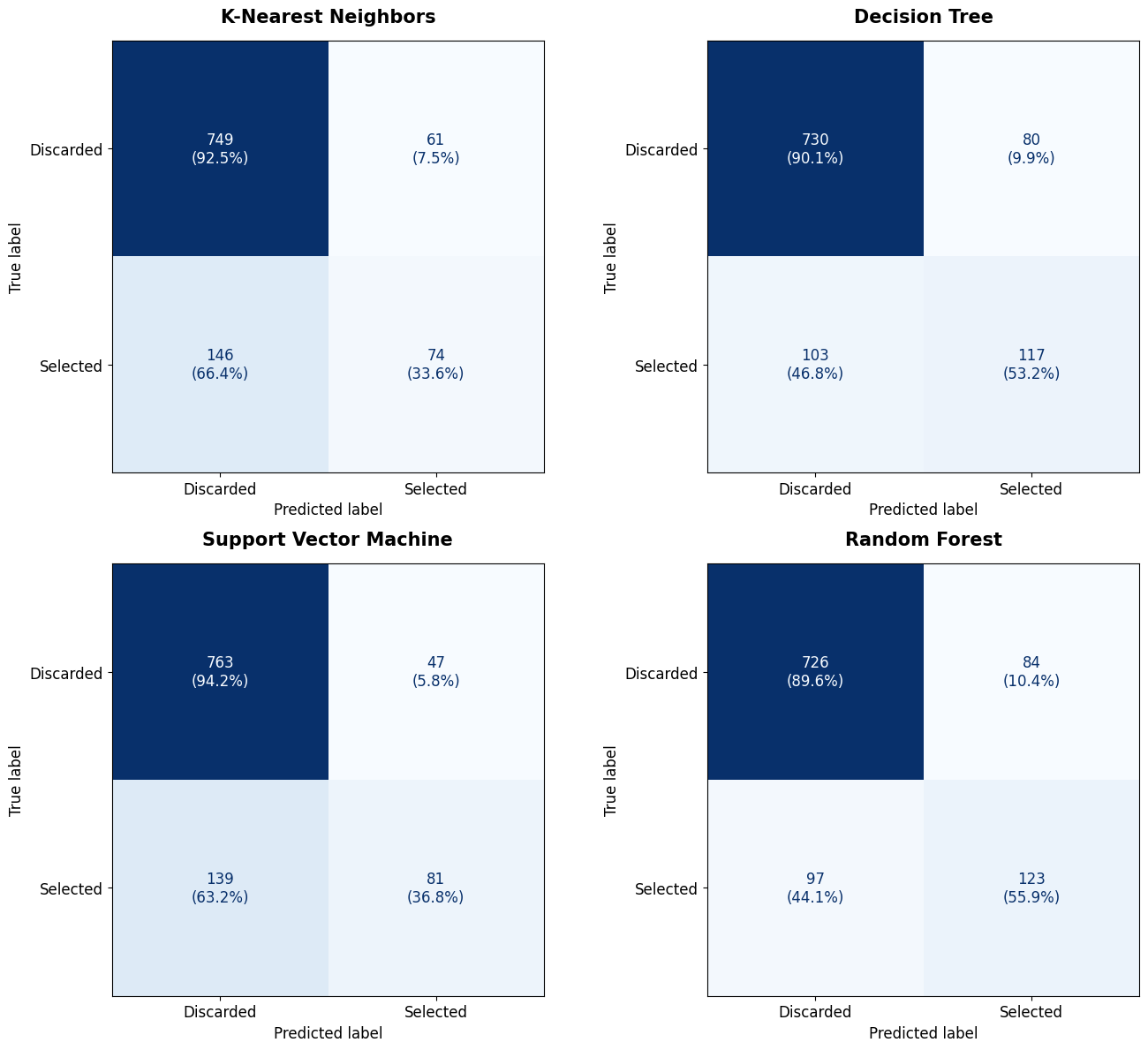}

    \vspace{0.6em}

    \includegraphics[width=0.6\columnwidth]{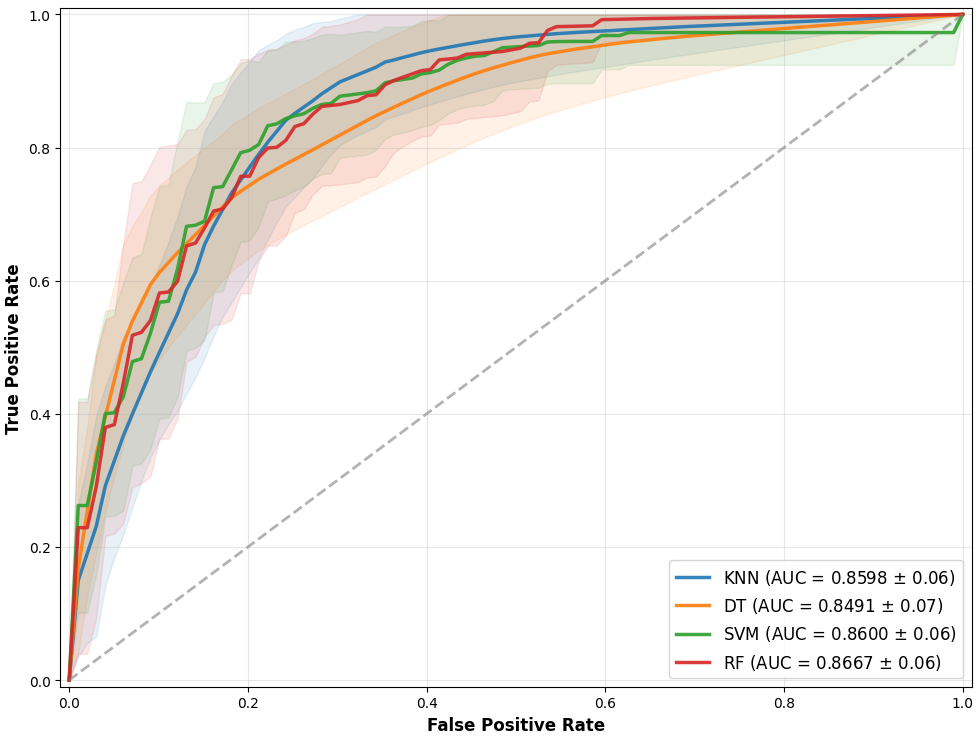}
    \caption{Top (a): Confusion matrix. Bottom (b): ROC-AUC of four models.}
    \label{f2}
\end{figure}

\textbf{Performance Comparison.}
Table~\ref{t1} reports the performance of four classifiers evaluated using nested cross-validation to identify an effective model for screening player-generated game levels. Performance was stable across inner and outer loops, indicating limited overfitting and reasonable generalization to unseen levels. Among the models tested, RF demonstrated the strongest overall performance, achieving the highest recall and F1-score while maintaining high overall discrimination. Confusion matrix results (Figure~\ref{f2}a) further show that RF was least likely to discard expert-validated levels, a critical requirement for preserving high-quality user-generated content in the proposed screening pipeline. Although SVM also achieved a strong ROC-AUC performance, it was more conservative in identifying selectable levels. Taken together, these results indicate that RF provides the most suitable balance of robustness and sensitivity for identifying high-quality player-generated game levels.
Figure~\ref{f2}b further compares the performance of the four classifiers. RF and SVM consistently consistently outperformed the other classifiers. SVM (86.22\%) demonstrated strong classification performance, whereas RF (86.57\%) achieves the best overall performance across four models.

\textbf{Feature Importance.}
To improve interpretability and reduce the dimensionality of the dataset, Lasso regression was used to identify the most informative features for predicting whether a player-generated level would be selected by experts. The analysis indicates that selection decisions are primarily associated with a small set of core structural features, including the number of player characters and goals, as well as the presence of one-way platforms, platform bubbles, and poppable bubbles. These features are central to gameplay mechanics, suggesting that expert judgments of level quality are closely tied to how mathematical actions and objectives are structured within a level rather than to peripheral or decorative elements.


\section{Discussion}
The present study identified the top actionable predictors of game features using the best-performing machine learning model, RF, out of all the classifiers compared and contrasted in this study. Of these predictors, player characters and goals emerged as the most critical elements. While these features represent fundamental game-based mechanics rather than explicit mathematical symbols, they serve as the essential interactive vessel for the learning content. The player character acts as the mathematical problem solver, whereas the goal defines the successful resolution of the game level. The findings demonstrate that the structural integrity of the game is a prerequisite for effective math learning during gameplay.
Also, features from physics objects groups play more important roles than features from obstacles, suggesting that the model prioritizes features that allow more constructive manipulation (related to calculation and logical reasoning) over passive barriers.
Our study uses machine learning methods to identify game features and predict selectable game levels, provide guidance for GBL systems in developing more intrinsic game levels. This automates the game level selection process and replaces labor-intensive manual curation. The proposed framework shows the possibility of using AI techniques to handle large volumes of player-generated game levels on the GBL systems and provide high-quality, intrinsically integrated learning experiences to players.

This study has several limitations. First, the size of the dataset constrains the model comparison. Future work includes collecting additional game levels to improve the robustness and generalizability of the results. Second,  the Creative Mode includes other game modules with different features. More predictors could be manually selected and explored for further study. Third, as these are preliminary findings, further exploration is needed and could focus on adaptive and personalized mathematical GBL informed by the proposed framework.
\section{ Conclusion}
This paper proposes a novel screening framework for mathematical GBL systems, applying machine learning methods to identify key game features and predict effective player-generated levels. Our findings rigorously demonstrate that RF is the optimal model for screening game levels in this case. By bridging a critical gap in the existing literature, our approach provides new insights for the development of mathematical GBL systems and enhances the creation of sufficiently intrinsic games. Furthermore, the ML classifier can maintain consistency and standardization in the screening process, ensuring the quality and reliability of game levels. Ultimately, this study provides a theoretical and practical foundation for the development of math learning games, also improving students' motivation, engagement, and critical thinking skills through gameplay.

\FloatBarrier

%

\bibliographystyle{splncs04}
\bibliography{references}

@article{dai2022narrative,
  title={Narrative-supported math problem solving in digital game-based learning},
  author={Dai, Chih-Pu and Ke, Fengfeng and Pan, Yanjun},
  journal={Educational technology research and development},
  volume={70},
  number={4},
  pages={1261--1281},
  year={2022},
  publisher={Springer}
}

@article{alam2025systematic,
  title={A Systematic Review and Meta-Analysis of Digital Mathematics Interventions for K-12 Students with Mathematical Learning Disabilities},
  author={Alam, Sabrina Shajeen and Gao, Jie and Dub{\'e}, Adam Kenneth},
  journal={International Journal of Science and Mathematics Education},
  pages={1--28},
  year={2025},
  publisher={Springer}
}

@book{liljedahl2016problem,
  title={Problem solving in mathematics education},
  author={Liljedahl, Peter and Santos-Trigo, Manuel and Malaspina, Uldarico and Bruder, Regina},
  year={2016},
  publisher={Springer Nature}
}

@article{ke2016designing,
  title={Designing and integrating purposeful learning in game play: A systematic review},
  author={Ke, Fengfeng},
  journal={Educational Technology Research and Development},
  volume={64},
  number={2},
  pages={219--244},
  year={2016},
  publisher={Springer}
}

@article{sharma2022game,
  title={Game Design for Mathematics Education},
  author={Sharma, Robin and Lajoie, Susanne P and Dub{\'e}, Adam Kenneth},
  journal={Mathematics Education: Research and Innovations},
  pages={25--37},
  year={2022}
}

@article{pan2023patterns,
  title={Patterns of using multimodal external representations in digital game-based learning},
  author={Pan, Yanjun and Ke, Fengfeng and Dai, Chih-Pu},
  journal={Journal of Educational Computing Research},
  volume={60},
  number={8},
  pages={1918--1941},
  year={2023},
  publisher={Sage Publications Sage CA: Los Angeles, CA}
}

@article{chiotaki2023adaptive,
  title={Adaptive game-based learning in education: a systematic review},
  author={Chiotaki, Domna and Poulopoulos, Vassilis and Karpouzis, Kostas},
  journal={Frontiers in Computer Science},
  volume={5},
  pages={1062350},
  year={2023},
  publisher={Frontiers Media SA}
}

@article{kacmaz2022examining,
  title={Examining pedagogical approaches and types of mathematics knowledge in educational games: A meta-analysis and critical review},
  author={Kacmaz, Gulsah and Dub{\'e}, Adam K},
  journal={Educational Research Review},
  volume={35},
  pages={100428},
  year={2022},
  publisher={Elsevier}
}

@inproceedings{mcewen2015engaging,
  title={Engaging or distracting: Children’s tablet computer use in education},
  author={McEwen, Rhonda N and Dube, Adam},
  year={2015},
  organization={International Forum of Educational Technology and Society}
}

@article{mayfield2002effects,
  title={The effects of cumulative practice on mathematics problem solving},
  author={Mayfield, Kristin H and Chase, Philip N},
  journal={Journal of applied behavior analysis},
  volume={35},
  number={2},
  pages={105--123},
  year={2002},
  publisher={Wiley Online Library}
}

@incollection{dube2016games,
  title={Are games a viable home numeracy practice?},
  author={Dub{\'e}, Adam K and Keenan, Andy},
  booktitle={Early childhood mathematics skill development in the home environment},
  pages={165--184},
  year={2016},
  publisher={Springer}
}

@article{konca2022digital,
  title={Digital technology usage of young children: Screen time and families},
  author={Konca, Ahmet Sami},
  journal={Early Childhood Education Journal},
  volume={50},
  number={7},
  pages={1097--1108},
  year={2022},
  publisher={Springer}
}

@article{cayton2015tablet,
  title={Tablet-based math assessment: What can we learn from math apps?},
  author={Cayton-Hodges, Gabrielle A and Feng, Gary and Pan, Xingyu},
  journal={Journal of Educational Technology \& Society},
  volume={18},
  number={2},
  pages={3--20},
  year={2015},
  publisher={JSTOR}
}

@article{mao2022effects,
  title={Effects of game-based learning on students’ critical thinking: A meta-analysis},
  author={Mao, Weijie and Cui, Yunhuo and Chiu, Ming M and Lei, Hao},
  journal={Journal of Educational Computing Research},
  volume={59},
  number={8},
  pages={1682--1708},
  year={2022},
  publisher={Sage Publications Sage CA: Los Angeles, CA}
}

@article{ninaus2017acceptance,
  title={Acceptance of game-based learning and intrinsic motivation as predictors for learning success and flow experience},
  author={Ninaus, Manuel and Moeller, Korbinian and McMullen, Jake and Kiili, Kristian},
  journal={International Journal of Serious Games},
  volume={4},
  number={3},
  pages={15--30},
  year={2017},
  publisher={Serious Games Society}
}

@article{chang2020experiencing,
  title={From experiencing to critical thinking: a contextual game-based learning approach to improving nursing students’ performance in electrocardiogram training},
  author={Chang, Ching-Yi and Kao, Chien-Huei and Hwang, Gwo-Jen and Lin, Fu-Huang},
  journal={Educational Technology Research and Development},
  volume={68},
  number={3},
  pages={1225--1245},
  year={2020},
  publisher={Springer}
}

@article{cicchino2015using,
  title={Using game-based learning to foster critical thinking in student discourse},
  author={Cicchino, Marc I},
  journal={Interdisciplinary Journal of Problem-Based Learning},
  volume={9},
  number={2},
  year={2015}
}

@article{hsiao2014development,
  title={Development of children's creativity and manual skills within digital game-based learning environment},
  author={Hsiao, H-S and Chang, C-S and Lin, C-Y and Hu, P-M},
  journal={Journal of Computer Assisted Learning},
  volume={30},
  number={4},
  pages={377--395},
  year={2014},
  publisher={Wiley Online Library}
}

@article{lee2024comparison,
  title={A comparison of machine learning algorithms for predicting student performance in an online mathematics game},
  author={Lee, Ji-Eun and Jindal, Amisha and Patki, Sanika Nitin and Gurung, Ashish and Norum, Reilly and Ottmar, Erin},
  journal={Interactive Learning Environments},
  volume={32},
  number={9},
  pages={5302--5316},
  year={2024},
  publisher={Taylor \& Francis}
}

@inproceedings{chi2014choice,
  title={Choice-based Assessment: Can Choices Made in Digital Games Predict 6th-Grade Students' Math Test Scores?},
  author={Chi, Min and Schwartz, Daniel and Blair, Kristen Pilner and Chin, Doris B},
  booktitle={Educational Data Mining 2014},
  year={2014}
}

@article{vandewaetere2013adaptivity,
  title={Adaptivity in educational games: Including player and gameplay characteristics.},
  author={Vandewaetere, Mieke and Cornillie, Frederik and Clarebout, Geraldine and Desmet, Piet},
  journal={International Journal of Higher Education},
  volume={2},
  number={2},
  pages={106--114},
  year={2013},
  publisher={ERIC}
}

\end{document}